\documentclass[twoside,11pt]{article}
\usepackage{jair, theapa, rawfonts}

\jairheading{1}{2017}{1-15}{10/2017}{--/--}
\ShortHeadings{Symbol Grounding Association in Multimodal Sequences with Missing Elements}
{Raue, Dengel, Breuel \& Liwicki}
\firstpageno{1}

\usepackage{graphicx}
\graphicspath{{figures/}}
\DeclareGraphicsExtensions{.png,.pdf}
\usepackage{algorithm}
\usepackage{algorithmic}
\usepackage{multirow}
\usepackage{todonotes}
\usepackage{changes}
\usepackage{bm}
\usepackage{booktabs}
\usepackage{amssymb}
\usepackage[caption=false]{subfig}
\usepackage{tikz}
\usepackage{color}
\usepackage{amsmath}
\usetikzlibrary{positioning}
\usetikzlibrary{arrows}
\usetikzlibrary{calc,shapes,shapes.geometric,automata,fit,backgrounds}
\begin{document}
\title{Symbol Grounding Association in Multimodal Sequences with Missing Elements}
\author{\name Federico Raue\textsuperscript{1,3} \email federico.raue@dfki.de\\
        \name Andreas Dengel\textsuperscript{1,3} \email andreas.dengel@dfki.de\\
        \name Thomas M. Breuel\textsuperscript{1} \email tmb@cs.uni-kl.de\\
        \name Marcus Liwicki\textsuperscript{2} \email liwicki@cs.uni-kl.de\\
        \addr \textsuperscript{1}Computer Science Department,TU Kaiserslautern, 
        Gottlieb-Daimler Strasse 1,\\
        67663 Kaiserslautern, Germany\\
        \addr \textsuperscript{2}MindGarage,TU Kaiserslautern, 
        Gottlieb-Daimler Strasse 1,\\
        67663 Kaiserslautern, Germany\\        
        \addr \textsuperscript{3}Smart Data and Knowledge Services,German Research Center for Artificial Intelligence (DFKI), 
        Trippstadter Strasse 122,\\
        67663 Kaiserslautern, Germany
        }
\date{October 2016}

\maketitle

\begin{abstract}
In this paper, we extend a symbolic association framework for being able to handle missing elements in multimodal sequences. The general scope of the work is the symbolic associations of object-word mappings as it happens in language development in infants. In other words, two different representations of the same abstract concepts can associate in both directions. This scenario has been long interested in Artificial Intelligence, Psychology, and Neuroscience. In this work, we extend a recent approach for multimodal sequences (visual and audio) to also cope with missing elements in one or both modalities. Our method uses two parallel Long Short-Term Memories (LSTMs) with a learning rule based on EM-algorithm. It aligns both LSTM outputs via Dynamic Time Warping (DTW). We propose to include an extra step for the combination with the max operation for exploiting the common elements between both sequences. The motivation behind is that the combination acts as a condition selector for choosing the best representation from both LSTMs. We evaluated the proposed extension in the following scenarios: missing elements in one modality (visual or audio) and missing elements in both modalities (visual and sound). The performance of our extension reaches better results than the original model and similar results to individual LSTM trained in each modality.
\end{abstract}
\maketitle

\section{Introduction}
\label{sec:Introducion}

A striking feature of the human brain is to link abstract concepts and sensory input signals, such as visual and audio.
As a result of this multimodal association, an abstract concept that has several representations (i.e., visual and audio) maps from one modality to another modality, and vice versa.
For example, the abstract concept ``ball'' from the sentence ``John  plays with a ball'' can be associated with several instances of different spherical shapes (visual input) and sound waves (audio input).
Several fields, such as Neuroscience, Psychology, and Artificial Intelligence, are interested in determining all factors that are involved in binding semantic concepts and the physical world.
This scenario is known as \emph{Symbol Grounding Problem}~\cite{harnad1990symbol} and is still an open problem~\cite{Steels2008}.

With this in mind, infants start learning the binding between abstract concepts and the real world in a multimodal scenario. \citeA{Gershkoff-Stowe2004} found the initial set of words in infants is mainly nouns, such as dad, mom, cat, and dog.
In contrast, the lack of stimulus can limit the language development~\cite{Andersen1993,Spencer2000}, i.e., deafness, blindness.
Asano \emph{et al.} found two different patterns in the brain activity of infants depending on the semantic correctness between a visual and an audio stimulus.
In simpler terms, the brain activity is pattern `A' if the visual and audio signals represent the same semantic concept.  Otherwise, the pattern is `B'.

Related work has been proposed, in which one model combines the \emph{Symbol Grounding Problem} and the association learning.
\citeA{yu2004multimodal} explored a framework that learns the association between objects and their spoken names in day-to-day tasks.
\shortciteA{nakamura2011grounding} introduced a multimodal categorization applied  to  robotics.
Their  framework  exploited  the  relation  of  concepts  in different modalities (visual, audio and haptic) using a Multimodal latent Dirichlet allocation.

Previous approaches have focused on associating isolated elements.
This work proposes another approach, where multimodal sequences are the input of the association task.
Text lines classification in OCR~\citeA{Breuel2013} and image segmentations~\citeA{byeon2015scene} are two successful examples of the sequence approach. 
Furthermore, the association task between sequences that can present  semantic concepts in one or two modalities.
Moreover, we are interested in multimodal sequences that represent a semantic concept sequence with the constraint that some elements that make up the concept are not part of both modalities; that is an element may be unique to one modality.
For instance, one modality sequence (text lines of digits) is represented by `2 4 6', and the other modality (spoken words) is represented by `two five six'; where `4' is unique to the text modality, and `five' is unique to the spoken modality.

In this work, we investigate the multimodal association of weakly labeled sequences based on the alignment between two \emph{latent spaces}.
In more detail, some elements of the sequence in one modality are not present in the other modality.
Note that our work is an extension of ~\shortciteA{RaueCoCo2015} where both modalities represent the same semantic sequence (no missing elements).
Similarly to \shortciteA{RaueCoCo2015}, two Long Short-Term Memories (LSTMs) are the main components of the presented model where the output vectors are aligned on the time axis using Dynamic Time Warping (DTW))~\cite{Berndt1994}.
Our contributions in this paper are the following

\begin{itemize}
    \item We propose a novel model for the cognitive multimodal association task (without claiming that being cognitively plausible).  Moreover, our model handles multimodal sequences where the semantic concepts can be in one or both modalities.  Also, a \emph{max operation} in the time-axis is novel in the architecture, and the motivation is to exploit the cross-modality of the shared semantic concepts.
    \item We evaluate the presented model in two scenarios.  In the first scenario, the missing semantic concepts can be in any modality.  In the second scenario, the semantic concepts are missing only in one modality.  For example, the visual sequence `1 2 3 4 5 6' and the audio sequence `two four'.  The visual and audio modalities share the semantic concepts \emph{two and four}.  In contrast, the semantic concepts \emph{one, three, five, and six} are presented only in the visual sequence.  In both cases, our model performances better than the model proposed by \shortciteA{RaueCoCo2015}.
\end{itemize}

This paper is organized as follows.
We shortly describe Long Short-Term Memory networks that are a recurrent neural network in Section~\ref{sub:LSTM}.
Section~\ref{sec:ctc} explains how LSTM can be trained based on weakly labeled sequences.
Section~\ref{sec:MSA} describes the original model for the object-word association.
Section~\ref{sec:MaxMean} presents the novel model for handling missing elements.
Section~\ref{sec:Experiment} shows a new dataset of multimodal sequences with missing elements.
In Section~\ref{sec:Result}, we compare the performance of the proposed extension, the original model, and a single LSTM network trained on one modality in the traditional setup (predefined coding scheme).  

\begin{figure}[t]
\begin{center}
\includegraphics[width=0.9\columnwidth]{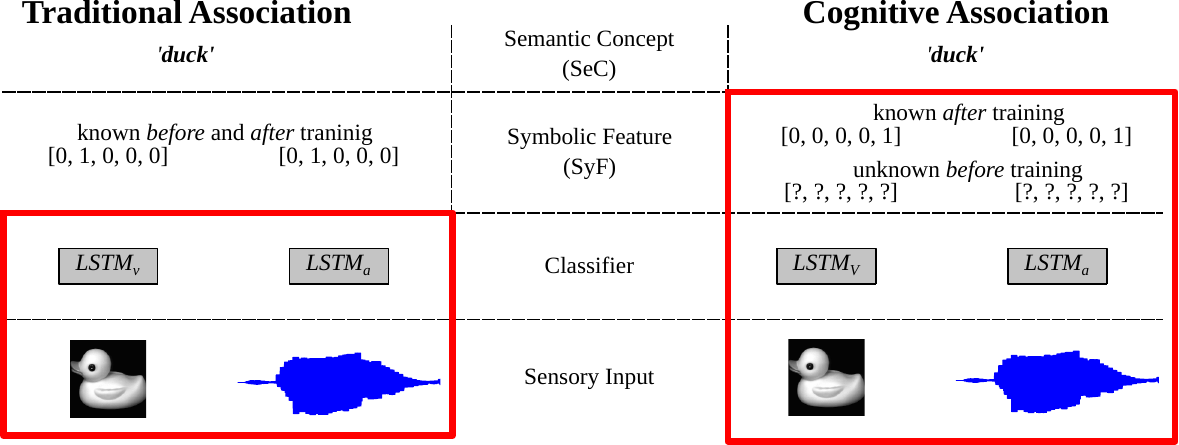}
\end{center}
\caption{Comparison of components between the traditional setup and our setup for associating two multimodal signals.  Note that our task has an extra learnable component (relation between semantic concepts and their representations), whereas the traditional scenario is already predefined (red box).  Moreover, the final goal is to agree on the same coding scheme for each modality.}
\label{fig:problem}
\end{figure}

\subsection{Symbol Grounding and Association Learning in Neural Networks}
In this work, we are interested in unifying the Symbol Grounding Problem and the association Learning in a neural network architecture.
For example, \citeA{dePenning2011neural} proposed a model that combines neural networks with temporal rules.
Consider the cross-modal scenario, in which an isolated object and a spoken word represent the same Semantic Concept.
One option for learning the association is to have two neural networks: one network for the visual channel and another for the audio channel.
It is safe to assume that the \emph{raw output vectors} of neural networks carry attributes or discriminant information of input samples.
Thus, the \emph{raw output vectors} can be considered as a \emph{numerical, symbolic feature}.
With this in mind, we can define the association task inspired by the Symbol Grounding Problem as follows.
The \emph{symbolic features} (raw output vectors) and semantic vectors are not binding initially.
This decision is usually taken before training and external to the network.
In contrast, the presented task requires that the model learn the binding.
Figure~\ref{fig:problem} shows a comparison between the \emph{traditional association task} and the presented task (inspired by the Symbol Grounding Problem).
It can be observed that the \emph{traditional association} has already defined two elements (red line - traditional association): a) the association is between two samples and b) the binding between semantic concepts and symbolic features.
Those elements are defined semantic concepts and the network output using the same one hot scheme.
On the other hand, the training algorithm includes the previous two elements (red line - cognitive association).
In this case, the network training incorporates two tasks.
First, each neural network learns the binding between semantic concepts and symbolic features.
Second, both neural networks are learning to agree to the same binding between symbolic features and semantic concepts.

\subsection{Multimodal Tasks in Machine Learning}
\label{subsec:ML}
Machine Learning has been applied successfully to several scenarios where the architecture exploits the multimodal relation between input samples.  In the following, we want to indicate the differences between previous multimodal tasks and our work.
\begin{description}
\item [Multimodal Feature Fusion] The task is to combine features of different modalities for creating a better feature.  In this manner, the generated feature exploits the best qualities of each modality.  Recently,  Deep Boltzmann Machines learns how to combine different modalities in unsupervised environments~\cite{srivastava2012multimodal,sohn2014improved}.  \shortciteA{iqbal2016scalable} proposed an architecture that combined three modalities, where the previous approaches combine only two modalities.

\item[Image Captioning] The task is to generate a textual description given images as input.  In other words, this task can be seen as a machine translation from images to captions.  One of the approaches to solving image captioning is a combination of Convolutional Neural Networks (CNN) and LSTM, where CNN encodes images and LSTM generates the textual descriptions~\cite{vinyals2014show,karpathy2014deep}.

\item[Weakly Labeled Association] This task is related to learn the association between two parallel sequences that represent the same order of semantic concepts.  One essential requirement is that both sequences are weakly labeled.  The two scenarios (image captioning and feature fusion) require that words must be already segmented  (more details in Section~\ref{sec:ctc}).  \shortciteA{RaueCoCo2015} proposed a model that has two parallel LSTM networks that exploit the multimodal latent space produced by LSTM networks. Both LSTMs align with each other for training.  The goal is that one LSTM of one modality learns with the latent space produced by the other modality (more details in Section~\ref{sec:MaxMean}).
\end{description}

\section{Long Short-Term Memory (LSTM)}
\label{sub:LSTM}

Long Short-Term Memory (LSTM) is a recurrent neural network that handles the vanishing gradient problem in long sequences~\cite{Hochreiter1997,Hochreiter1998}.
LSTM solves the vanishing gradient problem based on a set of gates that manages the flow information in three aspects: input, reset, and output. 
More formally, this architecture is defined by

\begin{equation}
\bm{i}_t = \sigma(\bm{W}_{xi}\bm{x}_t + \bm{W}_{hi}\bm{h}_{t-1} + \bm{b}_i)
\end{equation}
\begin{equation}
\bm{f}_t = \sigma(\bm{W}_{xf}\bm{x}_t + \bm{W}_{hf}\bm{h}_{t-1} + \bm{b}_f)
\end{equation}
\begin{equation}
\bm{o}_t = \sigma(\bm{W}_{xo}\bm{x}_t + \bm{W}_{ho}\bm{h}_{t-1} + \bm{b}_o)
\end{equation}
\begin{equation}
\bm{g}_t = \tanh(\bm{W}_{xc}\bm{x}_t + \bm{W}_{hc}\bm{h}_{t-1} + \bm{b}_c)
\end{equation}
\begin{equation}
\bm{c}_t = \bm{f}_t \bm{c}_{t-1} + \bm{i}_t \bm{g}_t 
\end{equation}
\begin{equation}
\bm{h}_t = \bm{o}_t \tanh(\bm{c}_t)
\end{equation}
\begin{equation}
\bm{z}_t = \text{softmax}(\bm{W}_{hz}\bm{h}_t + \bm{b}_z)
\end{equation}

where $\bm{x}_t \in \mathbb{R}^n$ is the input vector at time step $t$, $\bm{z}_t \in \mathbb{R}^c$ is the output vector at time step $t$, and $\bm{W}_{*}$ and $\bm{b}_{*}$ are the weight matrices and bias, respectively.

LSTM has trained in a similar way to any other gradient-based method.
Moreover, one of the most common approaches for training has been proposed by~\shortciteA{werbos1990backpropagation}.
He defined an algorithm, called \emph{Backpropagation Through Time (BPTT)}, which updates the network parameters from the last time step to the first time step.
In other words, the algorithm describes the loss function as follows

\begin{align}
J(t) &= \bm{y}_t - \bm{z}_t, \ \ t=T,\ldots, 1\\
\frac{\partial}{\partial \theta}  J &= \sum_{t=1}^{T} \frac{\partial }{\partial \theta} J(t)
\end{align}

where $\bm{y}_t \in \mathbb{R}^c$ is the target vector, $\theta$ is the network parameters, $\partial J(t)/\partial \theta$ is the derivative of the loss function w.r.t. to the network parameters.

So far, we have explained only one LSTM.
Additionally, another approach combines two LSTMs in the following manner.
One LSTM runs from 1 to T, and another LSTM runs from T to 1.
This setup is called \emph{Bidirectional LSTM}, and the motivation is to exploit the surrounded context of a specific position (i.e., before and after).

\section{Weakly Label Training for LSTM}
\label{sec:ctc}

\begin{figure}[t!]
\begin{center}
\input{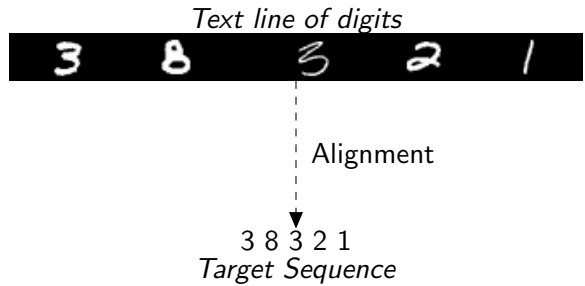}
\end{center}
\caption{Example of alignment between an input sequence and weakly labeled target sequence.  Note that the input sequence does not require to have a bounding box for each character.}
\label{fig:weakly_input_sample}
\end{figure}

LSTM has been successfully applied to several scenarios, such as image captioning~\cite{karpathy2014deep}, texture classification~\cite{byeon2015scene}, and machine translation~\cite{sutskever2014sequence}.
All previous examples require segmented data.
For example, words in sentences show a type of segmentation based on spaces (image captioning and machine translation tasks).
In this previous example, the segmentation of words is a relatively easy task.
In contrast, the segmentation in Speech Recognition and Optical Character Recognition (OCR) requires a vast human effort.
For example, consider the human effort of annotating bounding boxes for each character on this page.

\begin{figure}[t!]
\begin{center}
\input{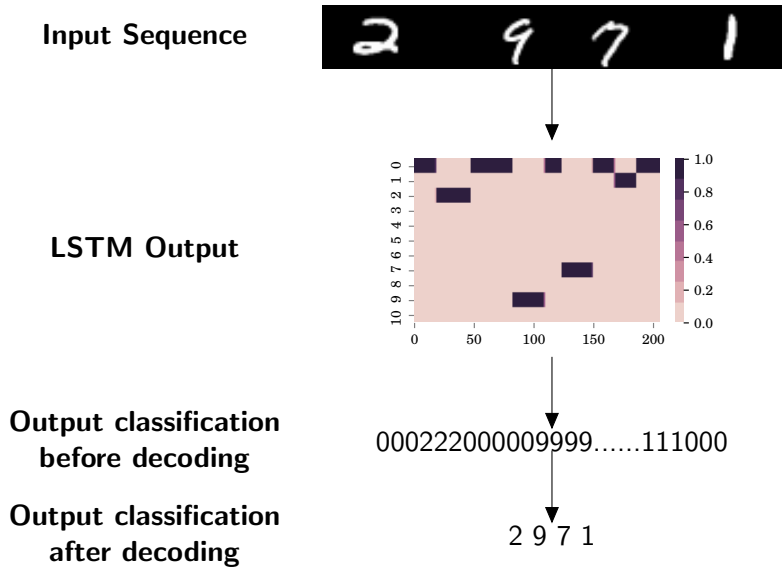}
\end{center}
\caption{Example of the LSTM classification based on CTC training.  The \emph{blank class} is represented by the index zero.  Note that the output classification is the same as the input sequence.}
\label{fig:CTC_example}
\end{figure}

In this work, we are interested in exploiting a training algorithm where LSTM can align an input sequence and target sequence.
For example, the input sequence can be a text line that represents the number ``38321'', and the target sequence is the string ``38321''.
Figure~\ref{fig:weakly_input_sample} shows an example of the weakly labeled sequence.
It can be observed the length of the input sequence is larger than the target sequence. 
\shortciteA{Graves2006} proposed \emph{Connectionist Temporal Classification (CTC)}.
The authors have included an extra class, called \emph{blank class (b)}.
So, the target sequence is re-written with the blank class "b3b8b3b2b1b".
The intuition of blank class is to learn the transition between digits (for example, from \emph{3} to \emph{8}) and to handle repeated characters (\emph{ll} from the word ``hello'').
After extending the target sequence, CTC layer exploits the similarities between LSTM and \emph{Hidden Markov Models (HMM)}.
In this case, LSTM uses a forward-backward procedure that is similar to HMM training algorithm.
CTC-forward-backward step requires two recursive variables forward ($fw$) and backward ($bw$) for generating the target vector $\bm{y}_{t}$. 
A \emph{forward-backward} algorithm employs the output vectors of LSTM.
The idea is to propagate the forward and backward the probabilities on the target sequence.
Finally, the output of the forward-backward algorithm is a target sequence for training LSTM.
The final step of the CTC training is to predict the label sequence given an unknown input sequence.  This step is called \emph{decoding}, and two methods have been proposed: Best Path Decoding and Prefix Search Decoding.
Figure~\ref{fig:CTC_example} shows an example of LSTM classification.
Please refer to the original paper for more 

\section{Multimodal Symbolic Association}
\label{sec:MSA}

The multimodal scenario is defined by a multimodal sequence (visual and audio).
Both channels represent the same sequence of semantic concepts and are defined by

\begin{align}
\bm{X}_{v}  =& \ \{ \bm{x}_{v,1}, \ldots, \bm{x}_{v,t1} \}, \\
\bm{X}_{a}  =& \ \{ \bm{x}_{a,1}, \ldots, \bm{x}_{a,t2} \},
\end{align}

where $\bm{x}_{v,*} \in \mathbb{R}^{n1}$ and $\bm{x}_{a,*} \in \mathbb{R}^{n2}$ are the input vectors for the (v)isual and (a)udio modalities (respectively).
The semantic concepts are represented by $SeC = \{s_1, \ldots, s_c\}$ where $c$ is the vocabulary size.
Thus, the ordered sequence for each input sequence is defined by 
\begin{align}
    SeC_v =& \ \{s_1, \ldots, s_k\}\\
    SeC_a =& \ \{s_1, \ldots, s_k\}
\end{align}

where $SeC_v \cap SeC_a = \{s_1, \ldots, s_k\}$.
The association proposed by~\shortciteA{RaueCoCo2015} exploits the benefits of CTC training where both sequences align to the same semantic sequence.
With this in mind, two bidirectional LSTM networks are defined for each modality $LSTM_v$ and $LSTM_a$.
In this manner, the training of $LSTM_v$ uses the latent space produced by $LSTM_a$, and vice versa.
The model  has two components.
The first component is to learn the binding between semantic concepts and symbolic features (Section \ref{sub:Statistical}).
The other component is the alignment between the multimodal space produced by both LSTMs (Section \ref{sub:DTW}).

In general, the model works as follows.
Initially, $LSTM_v$ and $LSTM_a$ receive as an input the visual and audio components of the multimodal sequence (respectively)
Consequently, each network produces the output vectors of both modalities after the last time step

\begin{align}
\bm{Z}_{v}  =& \ \{ \bm{z}_{v,1}, \ldots, \bm{z}_{v,t1} \}, \\
\bm{Z}_{a}  =& \ \{ \bm{z}_{a,1}, \ldots, \bm{z}_{a,t2} \},
\end{align}

where $\bm{z}_{v,*}$ and $\bm{z}_{a,*} \in \mathbb{R}^{c}$ are the output vectors.
At this point, each output sequence contributes for finding the most likely binding between semantic concepts and symbolic features.
Additionally, the indexes 4 or 10 (via one-hot coding vector) can represent the semantic concept \emph{duck}.
This decision is made internally by the model.
Therefore, two sets of \emph{concept vectors} are introduced to each modality $\gamma^v_i$ and $\gamma^a_i \in \mathbb{R}^{c}$, where $i=1, \ldots, c$.
The motivation behind is to implement a \emph{winning-take-all} rule where all semantic concepts and symbolic features have different binding relations between them (more details in Section~\ref{sub:Statistical}).
After determining the most likely binding, LSTM output of each modality and the binding are fed to CTC forward-backward step (\emph{c.f.} Section~\ref{sec:ctc}).
Consequently, the output of each modality are 

\begin{align}
\bm{Y}_{v}  =& \ \{ \bm{y}_{v,1}, \ldots, \bm{y}_{v,t1} \}, \\
\bm{Y}_{a}  =& \ \{ \bm{y}_{a,1}, \ldots, \bm{y}_{a,t2} \},
\end{align}

\begin{figure}[t!]
\begin{center}
\resizebox{\columnwidth}{!}{\input{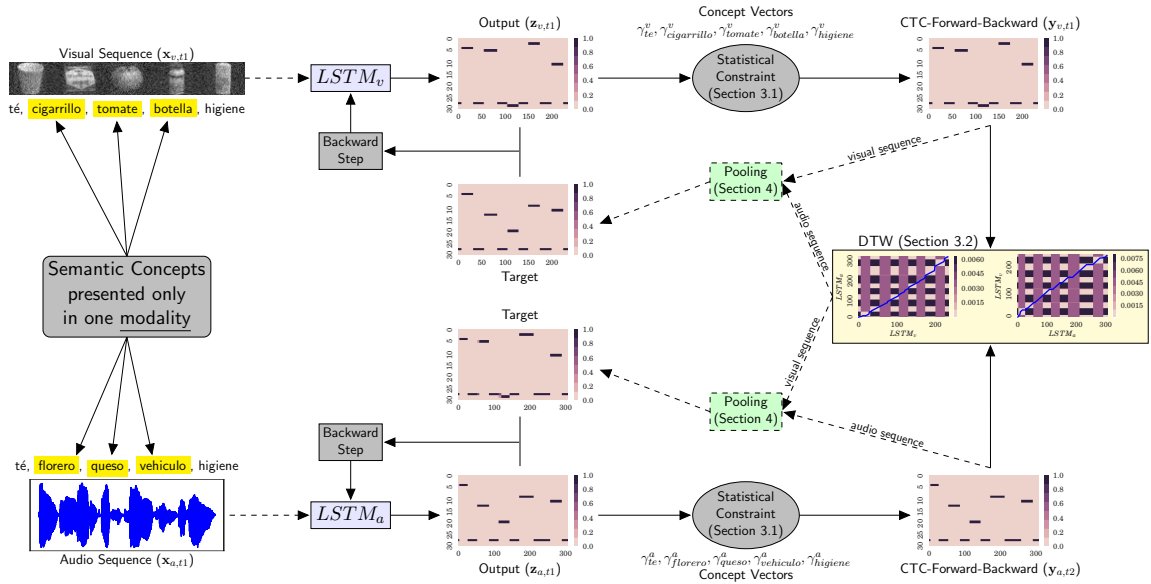}}
\end{center}
\caption{The new model can handle semantic concepts that are presented in one or two modalities.  In this work, we include a module that combines the audio and visual information based on the presence of semantic concepts.  Moreover, both channels are combined if they have the same semantic concept.Otherwise, there is no combination. Also, the \emph{max operation} improves the combination of the channels.}
\label{fig:Model}
\end{figure}

So far, the described steps are applied independently to each modality.
For learning the association and exploiting the multimodal latent space, both output vectors from CTC are aligned between them in the \emph{time-axis} by \emph{Dynamic Time Warping (DTW)}~\cite{Berndt1994}.
Hence, training on one modality uses the latent space of the other modality, and vice versa. 
Figure~\ref{fig:Model} illustrates the training algorithm with an example.

\begin{figure}[t!]
\begin{center}
\resizebox{\columnwidth}{!}{\input{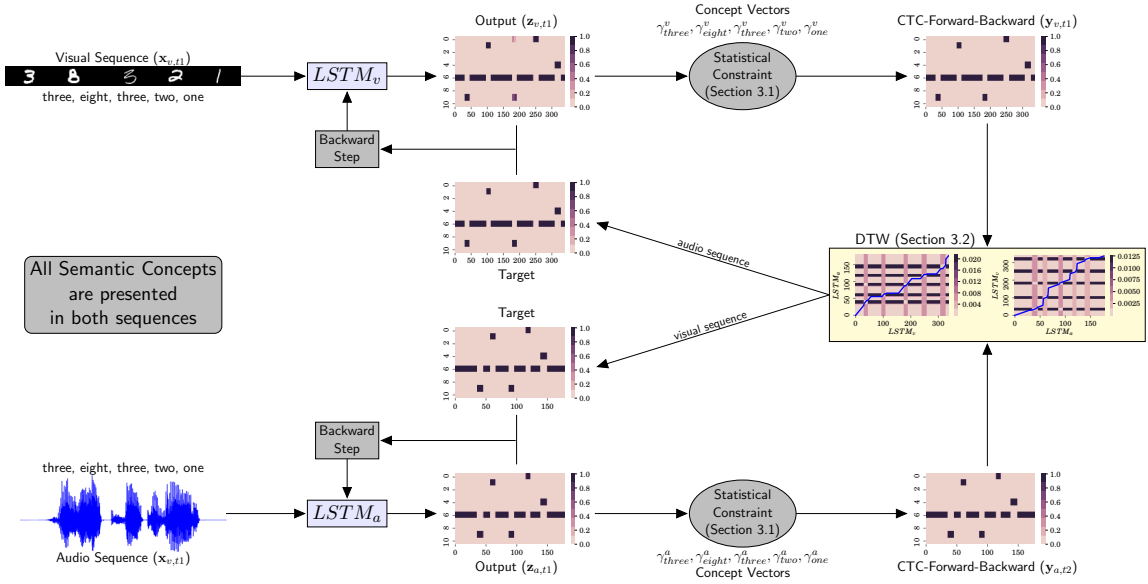}}
\end{center}
\caption{Association model based on two parallel bidirectional LSTM networks, which is proposed by~\shortciteA{RaueCoCo2015}. Note that each semantic concept is presented in both channels.  It can be observed that DTW module aligns the CTC layer produces by $LSTM_a$ to the CTC layer produces by $LSTM_v$.  As a result, the target vector for training $LSTM_v$ is purely obtained from $LSTM_a$.  The training of $LSTM_a$ based on $LSTM_v$ follows a similar process.}
\label{fig:Model_original}
\end{figure}

\subsection{Statistical Constraint for Semantic Binding}
\label{sub:Statistical}

In this association scenario, one crucial constraint is related to the binding between semantic concepts and symbolic features, which is not defined before training.
As mentioned, the binding between the semantic concepts and the output vectors is learned based a novel set of \emph{concept vectors}.
To be clear, note that two or more concepts cannot have the same vectorial representations.
This component employs an EM-style algorithm.  For explanation purposes, it is described considering only one LSTM and one set of \emph{concept vectors} $\gamma_1,\ldots,\gamma_c \in \mathbb{R}^{c}$.  However, it can be applied to two LSTM networks independently.  

The \emph{E-step} predicts the mapping between semantic concepts in the sequence and the symbolic representation given the LSTM output and the \emph{concept vectors}.  The first step is to combine the weighting vectors with the output sequence, which is defined by 

\begin{equation}
\bm{\hat{z}}_i = \frac{1}{T} \sum^{T}_{t=1} power\big(\bm{z}_t, \bm{\gamma}_i \big),\ \ \ i= 1,\ldots, c
\end{equation}

where $\bm{z}_t$ is the LSTM output vector at time $t$, $\bm{\gamma_i}$ is the concept vector, $T$ is the number of time steps of the sequence, and $power(\bm{z}_t, \bm{\gamma}_i)$ is the \emph{element-wise} power operation between the output vector $\bm{z}_t$ at time step $t$ and the concept vector $\bm{\gamma}_i$.
Then, a matrix is assembled by concatenating $\bm{\hat{z}}_1, \ldots, \bm{\hat{z}}_c$.
As a result, the assembled matrix represents the relation between semantic concepts (column) and the symbolic features (row).

\begin{equation}
\bm{\hat{Z}} = \big[\bm{\hat{z}}_1, \ldots, \bm{\hat{z}}_c \big]
\end{equation}
\begin{equation}
\bm{e}_1, \ldots,\bm{e}_c = g(\bm{\hat{Z}})
\end{equation}

where $g(\bm{\hat{Z}})$ is a row-column elimination, and  $\bm{e}_1, \ldots,\bm{e}_c \in \mathbb{R}^c$ are column vectors of a permutation of the identity matrix.  For simplicity, the column vector $\bm{e}_i$ can represent j-th identity vector where i and j can or cannot be the same.
In other words, the column vector $\bm{e}_4$ can represent the 1-st identity vector (e.g., $\bm{e}_4 = [1\ 0\ 0\ \ldots]^T$).
The row-column elimination procedure ranks all values in the matrix.
Next, the position (col, row), where the maximum value is found and determines the \emph{row-th} identity column vector $\bm{e}_{col}$.
For example, the maximum value is found at (2, 5), and its correspondence vector is $\bm{e}_2 = [0\ 0\ 0\ 0\ 1\ 0\ \ldots]^T$.
All values of the previous column and the previous row are set to zero.
This column-row elimination is applied $c$ times.
Hence, the vectors $\bm{e}_1, \ldots,\bm{e}_c$ are the mapping between semantic concepts (columns) and their symbolic feature (rows).  

The \emph{M-step} updates the concept vectors given the LSTM output and the statistical distribution target.  Hence, the cost function is defined by 

\begin{equation}
cost_i = \bigg(\bm{\hat{z}_i} - \frac{1}{c} \bm{e}_i \bigg)^2, \ \ \ i=1, \ldots, c
\end{equation}
\begin{equation}
\bm{\gamma}_i = \bm{\gamma}_i - \alpha * \nabla_{\bm{\gamma}_i} cost_i, \ \ \ i=1, \ldots, c
\end{equation}

where $\bm{e}_i$ is a column vector of the identity matrix that represents the semantic concept, $\alpha$ is the learning rate, and $\nabla_{\bm{\gamma}_i} cost_i$ is the derivative w.r.t $\bm{\gamma}_i$.

\subsection{Dynamic Time Warping (DTW)}
\label{sub:DTW}

In this module, the goal is to combine both LSTMs in the latent space.
This combination is possible because the multimodal sequence represents the same sequence of semantic concepts and the monotonic behavior of LSTM.
The alignment is between CTC output of both modalities.
Moreover, \shortciteA{Berndt1994} proposed DTW for aligning two signals.
Similarly, DTW can also utilize LSTM output sequences.
DTW requires two steps for alignment of two signals. 
The first step is to calculate a cost matrix with the following relation

\begin{equation}
\label{eq:dtw}
DTW[i,j] = dist[i,j] + min \left\{\begin{array}{l}
   DTW[i-1,j-1]\\
   DTW[i-1,j] \\
   DTW[i,j-1]
  \end{array} \right.
\end{equation}

where $dist[i,j]$ is the \emph{Euclidean} distance between output vectors at timestep $i$ of $LSTM_v$ and at timestep $j$ of $LSTM_a$. 
The Second step is the alignment path between both LSTMs.
In this case, the path is a set of tuples that maps one time step of one LSTM to another time step of the other LSTM.
In other words, there is a function $f(s1, s2): s1 \rightarrow s2$, where $s1$ is the source sequence and $s2$ is the target sequence.
Afterwards, the loss function of one modality can use the other modality as a target, and vice versa.
Hence, the losses are defined by

\begin{equation}
J_v(t) = f_{a\rightarrow v}\left(\bm{y}_{a,t^{'}}\right)_t - \bm{z}_{v,t},\ \ \ \mbox{where $t=1,\ldots, t1$}
\label{eq:Orig1}
\end{equation}

\begin{equation}
J_a(t) = f_{v\rightarrow a}\left(\bm{y}_{v,t^{'}}\right)_t - \bm{z}_{a,t},\ \ \ \mbox{where $t=1,\ldots, t2$}
\label{eq:Orig2}
\end{equation}

where $f_{a\rightarrow v}$ and $f_{v\rightarrow a}$ are the alignment path from audio to visual and from visual to audio (respectively), $\bm{z}_{v,t}$ and $\bm{z}_{a,t}$ are LSTM output vector at time $t$, $\bm{y}_{v,t^{'}}$ and $\bm{y}_{a,t^{'}}$ are the CTC-forward-backward steps.  Note that $t^{'}$ is the time step in the source modality and time step $t$ is the time step in the target modality. 

\section{Handling Missing Elements}
\label{sec:MaxMean}

The previous section described a multimodal association model that exploits weakly labeled samples based on CTC training.
The initial assumption is that each channel in the multimodal sequence represents the same ordered sequence of semantic concepts.
However, the initial assumption is a limitation.
In this paper we present, an extension that handles missing elements in the sequences.
The goal is to exploit semantic concepts that are presented in both modalities.
Additionally, the multimodal combination can be boosted using a \emph{max operation} that combines the best of each modality, whereas the previous model only exploits one modality.  
More formally, the association task can be rewritten as follows
\begin{align}
\bm{X}_{v}  =& \ \{ \bm{x}_{v,1}, \ldots, \bm{x}_{v,t1} \}, \\
\bm{X}_{a}  =& \ \{ \bm{x}_{a,1}, \ldots, \bm{x}_{a,t2} \},\\
SeC_v =& \ \{s_1, \ldots, s_{k1}\},\\
SeC_a =& \ \{s_1, \ldots, s_{k2}\}
\end{align}
where $SeC_v \cap SeC_a  \neq \{ \emptyset \}$ and $|SeC_v \cap SeC_a| \leq \text{min}(|SeC_v|,|SeC_a|)$.

The coupling both LSTMs occurs in DTW module.
Similar to Section~\ref{sec:MSA}, this model follows the same procedure until DTW step.
As a reminder, the output of DTW is an alignment function from one modality to the other modality.
In other words, a function $f(s1, s2): s1 \rightarrow s2$ that maps the closest signal to $s2$ (target) based on the values of $s1$ (source).

The two main differences with the model proposed by \shortciteA{RaueCoCo2015} relies on two parts.
The first part is to combine only semantic concepts that are shared between both channels because the probability distribution produces by LSTMs should be similar (even if different input feature space produces them).
Furthermore, the max operation is a common approach for combining two vectors.
The second part is related to the semantic concepts that are presented only in one channel.
In this case, it is not required to use another modality.
Thus, this step can be formulated as follows

\begin{align}
\mathbf{\hat{y}}_{a\rightarrow v,t} =& \left\{
  \begin{array}{ll}
    \text{max}\left(y^k_{a,t^{'}}, y^k_{v,t}\right) & \mbox{\small semantic concept $k$ is presented in both modalities}\\
    y^k_{v,t} & \mbox{\small otherwise}
  \end{array}
\right.
\\
\mathbf{\hat{y}}_{v\rightarrow a,t} =& \left\{
  \begin{array}{ll}
    \text{max}\left(y^k_{v,t^{'}}, y^k_{a,t}\right) & \mbox{\small semantic concept $k$ is presented in both modalities}\\
    y^k_{a,t} & \mbox{\small otherwise}
  \end{array}
\right.
\end{align}

where $y^k_{v,t}$ and $y^k_{a,t}$ are the probability of semantic concept $k$ at time step $t$, $y^k_{v,t^{'}}$ and $y^k_{a,t^{'}}$ are the latent space produced by CTC layer for each modality, and function $max$ is the \emph{element-wise} maximum operation.
Note that $y^k_{v,t}, y^k_{a,t}, y^k_{v,t^{'}}$, and $y^k_{a,t^{'}}$ are scalar, and the goal is to assemble the vectors $\mathbf{\hat{y}}_{a\rightarrow v,t}$ and $\mathbf{\hat{y}}_{v\rightarrow a,t}$ that combines both cases (shared and non-shared semantic concepts).
Afterwards, the target vectors are used based on this operation.
The Equations \ref{eq:Orig1} and \ref{eq:Orig2} are updated 

\begin{align}
J_v(t) =&\ \ \mathbf{\hat{y}}_{a\rightarrow v,t} - \bm{z}_{v,t}, \ \mbox{where $t=1,\ldots, t1$} \label{eq:pool1}\\ 
J_a(t) =&\ \  \mathbf{\hat{y}}_{v\rightarrow a,t} - \bm{z}_{a,t}, \ \mbox{where $t=1,\ldots, t2$} \label{eq:pool2}
\end{align}

\section{Experimental Design}
\label{sec:Experiment}
\subsection{Datasets}
We generated several multimodal datasets where the elements of the sequence are missing in one or both modalities, but the relative order between the elements is the same.
For example, a visual semantic concept sequence is a text line of digits ``2 4 7'', and an audio semantic concept sequence can be represented by ``two seven''.
In this case, we assumed a simplified scenario of symbol grounding, where the continuity of semantic concepts is different on each modality.
The visual component is a horizontal arrangement of isolated objects, and the audio component is spoken semantic concepts of some elements of the visual component, and vice versa.  We want to point out that the visual component is similar to a panorama view.
The procedure for generating the multimodal datasets is explained.

\textbf{Generating Semantic Sequences:} Two scenarios are considered for generating the semantic sequences for each modality: missing elements in both modalities and one modality.  For the first scenario, we generated a sequence of ten semantic concepts.  Later,  we randomly remove between zero and five elements from each sequence.  As a result, two different sequences for two different modalities were obtained with few common elements between them.  For the second scenario, we follow a similar procedure.  In that case, one modality has a sequence with ten semantic concepts, and the other modality has missing elements.  For example, $|S_a|=10$ and $|S_v|= S_a - \text{missing elements}$.  In addition, our vocabulary has 30 semantic concepts in Spanish: \emph{oso, bote, botella, bol, caja, carro, gato, queso, cigarrillo, gaseosa, bebida, pato, cara, comida, hamburguesa, higiene, liquido, loci\'on, cebolla, piment\'on, pera, redondo, sanduche, cuchara, t\'e, tel\'efono, tomate, florero, veh\'iculo, madera}.

\textbf{Visual Component:}  We used a subset of 30 objects from COIL-100~\cite{Nene1996} that is a standard dataset of 100 isolated objects.  Each isolated object has 72 views at different angles with a black background.  After selecting the object for the sequence, each object was converted to grayscale and re-scaled to 32 x 32 pixels.  The visual components are composed by horizontally stacking isolated objects.  Additionally, the final image has been added a random noise as a background.  In this way, the segmentation step is more challenging with relation to a clean black background.  While the training set contains images of odd angles, the testing set has images of even angles.

\textbf{Audio Component:} We recorded each semantic concept two times from twelve different subjects who are Spanish native speakers (five female and seven male speakers) from different countries of Center and South America.  Afterwards, concatenating isolated semantic concepts generates the audio sequences. The training set contains eleven voices for training, whereas the testing set contains two.  

\begin{figure}[t]
\begin{center}
\includegraphics[width=0.9\columnwidth]{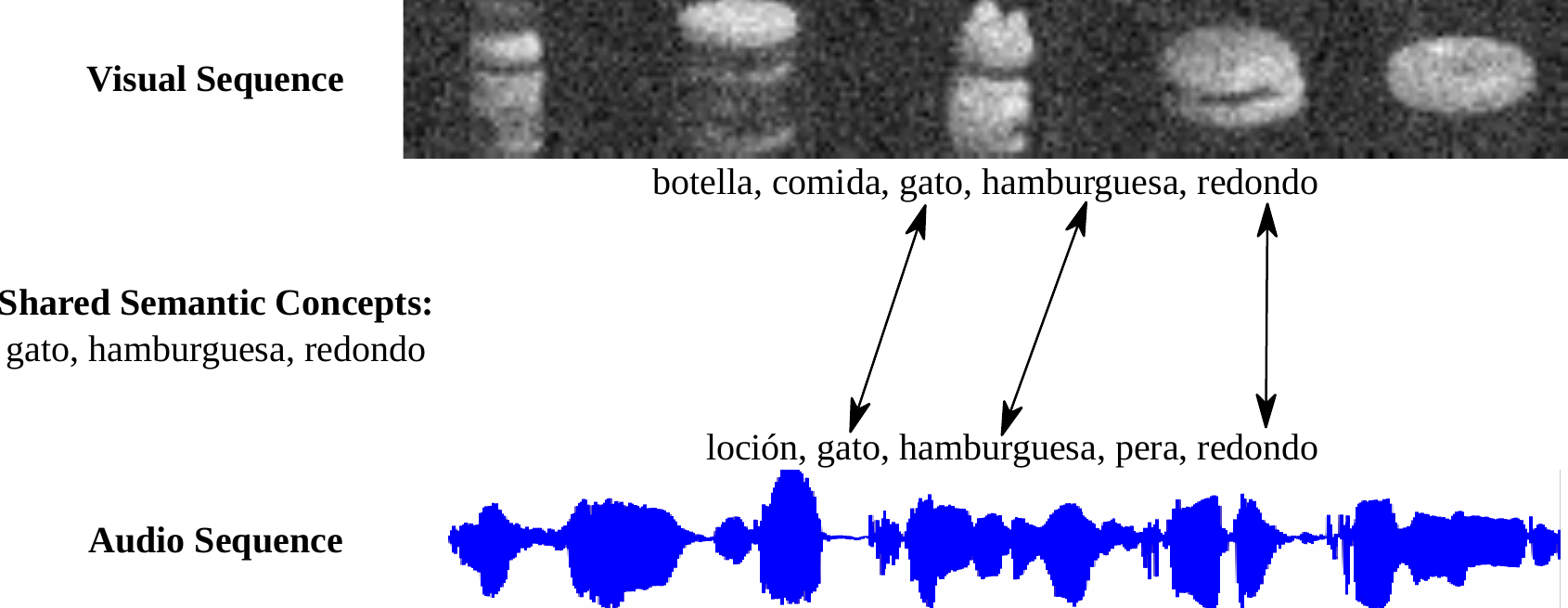}
\end{center}
\caption{Example of the multimodal dataset.  It can be observed that only three elements are the same on both modalities.}
\label{fig:Sample}
\end{figure}

\textbf{Training and Testing Multimodal Datasets:} We have generated three different multimodal configurations for evaluating our model.  The first setup has missing elements in both modalities.  While the second setup has ten semantic concepts in the visual component, the audio modality has a fixed number of missing elements.  Thus, we can evaluate the impact of the missing elements.  This setup covers from zero to five missing semantic concepts.  The third dataset has a similar idea concerning the second dataset.  In this case, we are testing an audio sequence with ten semantic concepts, but the visual component has a fixed number of missing elements. 
Additionally, 1,000 multimodal sequences have been generated for each subject in each setup.
We follow a 5-fold cross-validation scheme where eleven subjects are selected for training and the remaining two subjects are used for testing.
One example with both elements missing from both modalities is shown in Figure~\ref{fig:Sample}. 

\subsection{Input Features and LSTM setup}
We did not apply any pre-processing step for the visual component. 
In contrast, the audio component was converted to Mel-Frequency Cepstral Coefficient (MFCC) using HTK toolkit\footnote{http://htk.eng.cam.ac.uk}.  The audio representation is a vector of 123 components: a Fourier filter-bank with 40 coefficients (plus energy), including the first and second derivatives.  All audio and visual components were normalized to have zero mean and standard deviation one.

Also, the proposed extension was compared against the original model in \cite{RaueCoCo2015}.  Also, we compared the extension against LSTM with CTC layer and a predefined coding scheme.  The parameters of the visual LSTM were: 40 memory cells, learning rate 0.0001, and momentum 0.9.  On the other hand, the audio LSTM had 100 memory cells, and the learning rate and momentum are the same as in the visual LSTM.  Furthermore, the learning rate in the statistical constraint was set to 0.001.  The parameters are selected based on the best performance of LSTM that is trained independently on each modality.  

\section{Results and Discussion}
\label{sec:Result}
As mentioned previously, the assumption of the original model was to represent the same semantic concept sequence in both modalities.
In other words, a one-to-one relationship exists between modalities.  
In contrast, in this work, our assumption is more challenging because the semantic concept in one modality can be or cannot be present in the other modality.  We evaluate the multimodal association task using \emph{Association Accuracy (AAcc)}, which is defined by the following equation 

\begin{equation}
\centering
 AAcc = \frac{\sum_{i=1}^{N} LCS(output_{a,i}, output_{v,i}, gt_{a,i}, gt_{v,i})}{\sum_{i=1}^{N} LCS(gt_{a,i}, gt_{v,i})},
\end{equation}

where $LCS$ is the length of the longest common sequence, $output_{a,i}$ and $output_{v,i}$ are the output classification of each modality, $gt_{a,i}$ and $gt_{v,i}$ are the ground-truth labels of each modality, and $N$ is the number of elements in the dataset.  In other words, we are evaluating the association between the common elements.  Our model not only learns the association but also learns to classify each modality.  With this in mind, we also reported the Label Error Rate (LER) as a performance metric, which is defined by

\begin{equation}
LER = \frac{1}{N} \sum_{i=1}^N  \frac{ED(output_i, gt_i)}{|gt_i|}
\end{equation}

where $output_i$ is the output classification, $gt_i$ is the ground-truth, and $ED(output_i, gt_i)$ is the edit distance between the output classification and the ground-truth.  As a reminder, the training set has 11,000 multimodal sequences, whereas the testing set has 2,000 multimodal sequences.
In this work, we have reported the average results.

\begin{table}[t]
\centering
\caption{Association Accuracy (\%) and Label Error Rate (\%) from the multimodal dataset that has missing elements in both modalities.  It can be seen that the original model performs worse than the proposed combination.  Furthermore, the presented extension reached similar results to LSTM (trained for the easier classification task, and not for association) and under some conditions reaches better results (audio component).}
\label{tab:results}
\begin{tabular}{lccc}
\toprule
Model                 & Association & \multicolumn{2}{c}{Label Error Rate (\%)} \\ \cline{3-4}
                         &   Accuracy (\%)                         & visual              & audio               \\
\midrule
LSTM + CTC (baseline)                               & $70.68 \pm\  6.12$            & $0.14 \pm 0.14$       & $35.84 \pm \ 5.35$  \\  
Original Model (\shortciteA{RaueCoCo2015})          & $19.59 \pm\  8.90$            & $7.00 \pm 2.42$       & $79.01 \pm \ 9.51$      \\
\textbf{Our work}                                   & $71.52 \pm 11.85$            & $0.97 \pm 1.58$       & $33.09 \pm 10.38$     \\
   
\bottomrule
\end{tabular}
\end{table}

Table~\ref{tab:results} summarizes the performance of LSTM trained with a predefined coding scheme, the original model, and the presented extension.
Those results are divided into two parts as follows.
First, the proposed extension handles missing elements in multimodal sequences better than the original model.
It can be inferred that the max operation keeps the strongest of the common semantic concepts between modalities.
Note that the representations update the weights in the backward step.
Second, the proposed extension reaches similar results to the standard LSTM.
In this case, LSTM was trained in each modality independently.
As a reminder, we mentioned two setups for classification tasks: the traditional setup and the setup used in this work.
We want to point out that the visual LSTM boost the performance of audio sequences compared to LSTM.  As a result, our model reaches lower Label Error Rate in the audio sequences than the standard LSTM trained only in audio sequence.

\begin{figure}[t]
\begin{center}
\resizebox{\columnwidth}{!}{\input{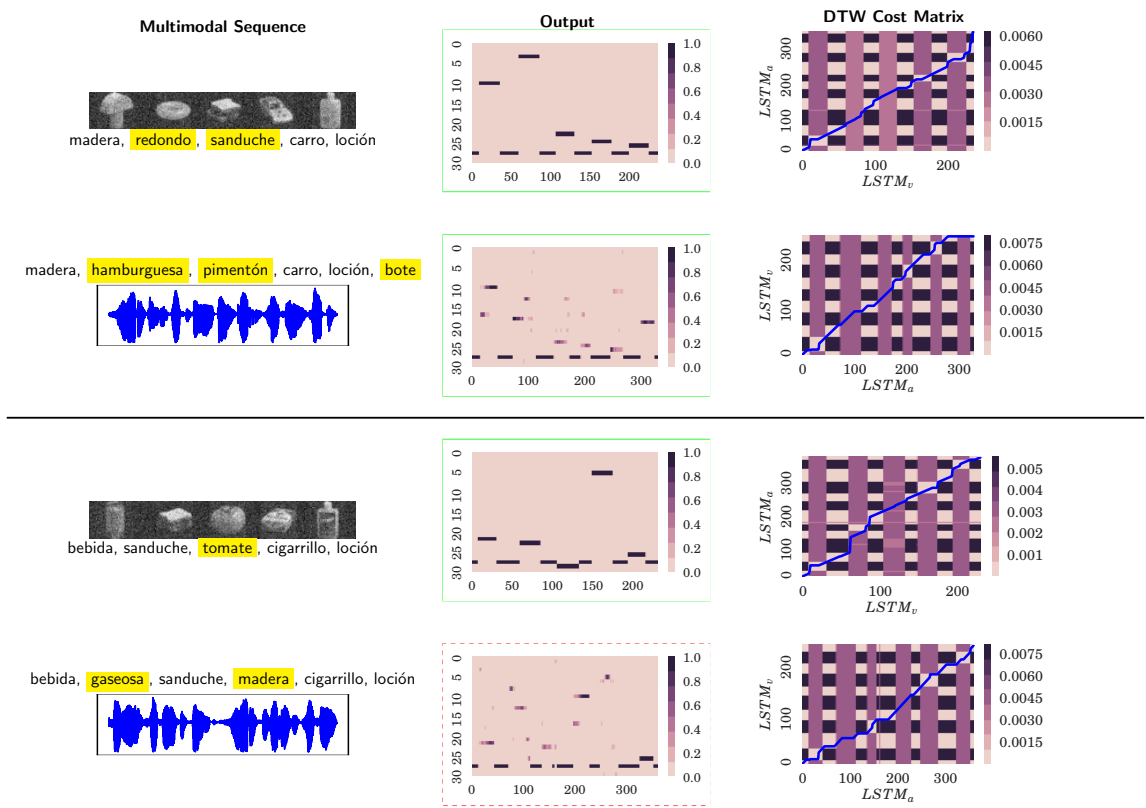}}
\end{center}
\caption{Several examples of the output classification and DTW cost matrices.  The first multimodal sequence shows an example, in which both output classifications are correct (solid green square).  The second multimodal sequence shows one correct and one incorrect output classification (dashed red square).}
\label{fig:DTW example}
\end{figure}

Another outcome of this work is the conformity of the symbolic structure in both modalities, even with missing elements.  
Figure~\ref{fig:DTW example} shows examples of the coding scheme agreement.
It can be observed that both LSTM networks learn to classify the object-word relation in weakly labeled multimodal sequences.
Moreover, the common concepts in both modalities are represented by a similar symbolic feature and located at the right position in the sequence.
For example, the semantic concept ``madera'' (first element at the visual and audio components) is represented by the index ``10'' in both modalities\footnote{There are some cases that represent one semantic concept with two different coding vectors for each network.  However, both networks retrieve correctly the same concept regardless of the different coding scheme.}.  Note that not only the common elements but also the missing elements are classified correctly.  Furthermore, the common semantic concepts are correctly classified  even if LSTM does not correctly classify  all semantic concepts of the sequence.  The second example in Figure~\ref{fig:DTW example} shows the semantic concept ``loci\'on'' is represented by index ``25''. 

\begin{figure}[t]
\begin{center}
\includegraphics[width=\textwidth]{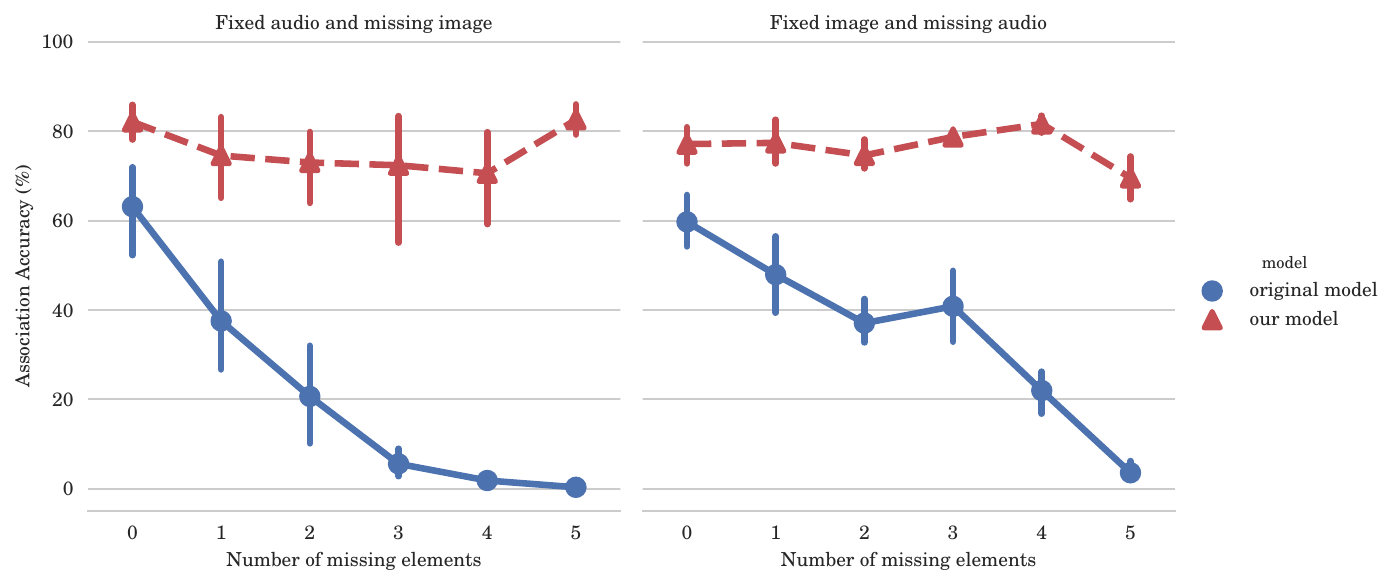}
\end{center}
\caption{Association of two multimodal setups.  In this case, one modality has ten semantic concepts, and the other modality has a fixed number of missing elements.  The presented model (triangle) outperforms to the original model (circle) regardless of the modality and number of missing elements.}
\label{fig:Comparison}
\end{figure}

\begin{figure}[t]
\begin{center}
\includegraphics[width=\textwidth]{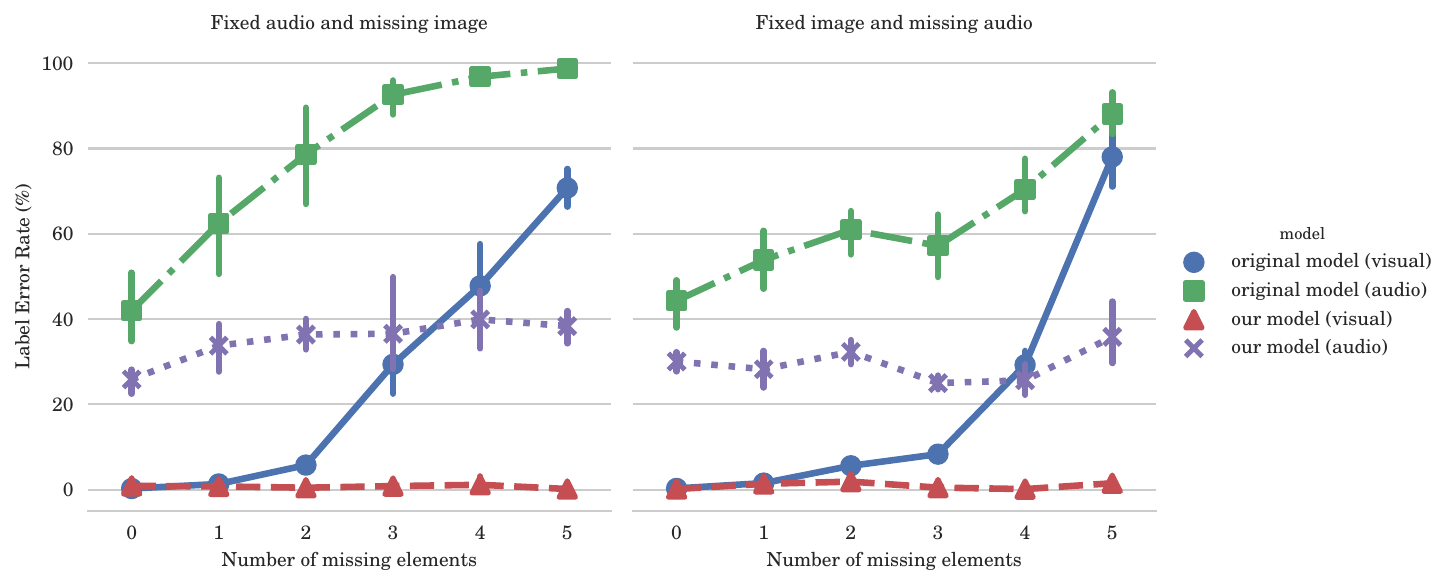}
\end{center}
\caption{Label Error Rate (\%) of each modality for each multimodal setup.  It can be observed that the presented model (visual-triangle and audio-X) keeps similar performances across the number of missing elements.  In contrast, the original model (visual-circle and audio-square) increases the error w.r.t. the number of missing elements.}
\label{fig:Comparison2}
\end{figure}

In addition to the considerations we made so far, we were also interested in the robustness of the presented model against the number of missing elements.  With this in mind, we generated several datasets where one modality has ten semantic concepts, and the other has only a fixed number of missing elements from the ten semantic concepts.  Figure~\ref{fig:Comparison} shows the \emph{Association Accuracy} of the original model and the presented model for handling missing elements.  First, the original model (solid blue line) decreases its performance when the number of missing elements increases in both modalities.  These results were expected because the original model relies on \emph{the one-to-one relation} between modalities.  Second, we recognize that the presented model (dashed red line) shows a better performance compared to the original model (solid blue line) in both modalities.  Thus, we may conclude that the presented model does not reduce its performance even if 50\% of elements are missing in one of the modalities.
Note that the max operation boosts the performance of our model when there are zero missing elements.

Figure~\ref{fig:Comparison2} shows that Label Error Rate of each modality.
One pattern appears at zero missing elements.
Our model reaches better performance than the original model because of the audio modality.
In this case, the combination of audio and visual latent spaces helps to reduce the error.
Moreover, it can be observed that the error of our model does not increase the same rate as the original model.

\section{Conclusions}
In summary, we have presented a solution inspired by the \emph{symbol grounding problem} for the object-word association problem.
Additionally, the model relies on multimodal sequences (visual and audio) where the semantic elements can be presented in one or both modalities.
However, we believe that one interesting direction is to analyze more quantitative if the model is cognitive plausible.
Further work is planned for more realistic scenarios where the visual component is not segmentable.
Moreover, we are interested in extending the word-association problem between a two-dimensional image and speech.
With this in mind, we will incorporate visual attention mechanism in synchronization with speech.
In the future, the model will be evaluated to add more modalities, for instance visual, audio, and motor sensors, the three modalities can be aligned between them.
Note that each sensorial input collects data of the same action.
Two approaches can be considered for the alignment step.
One option is to apply DTW for three dimensions (similar to \shortciteA{wollmer2009multidimensional}).
The other approach is to align between each pair signals and evaluate the most suitable relation.
For example, visual-audio, motor-visual, and audio-motor.
Finally, the human language development relies on the relationship between abstract concepts and the real world collected by the sensory input. 
The scenario of the symbol grounding problem might be considered as simple.
However, many questions remain still open~\cite{needham2005protocols,Steels2008}.


\vskip 0.2in
\bibliography{nips2016_conference}
\bibliographystyle{theapa}

\end{document}